\documentclass[conference]{IEEEtran}

\IEEEoverridecommandlockouts
\usepackage{cite}
\usepackage{amsmath,amssymb,amsfonts}
\usepackage{algorithmic}
\usepackage{graphicx}
\usepackage{textcomp}
\usepackage{multirow}
\usepackage{subcaption}
\usepackage{makecell}
\usepackage{booktabs}
\usepackage[table]{xcolor}% http://ctan.org/pkg/xcolor

\newcommand{\deltup}[1]{\ensuremath{\uparrow}\textcolor{red}{#1}}
\newcommand{\deltdown}[1]{\ensuremath{\downarrow}\textcolor{green}{#1}}
\definecolor{awesome}{rgb}{1.0, 0.13, 0.32}

\usepackage[colorlinks,linkcolor=blue,anchorcolor=blue,urlcolor=magenta,citecolor=awesome]{hyperref}
\def\BibTeX{{\rm B\kern-.05em{\sc i\kern-.025em b}\kern-.08em
    T\kern-.1667em\lower.7ex\hbox{E}\kern-.125emX}}
\begin{document}

\title{Exploring Self-supervised Skeleton-based Action Recognition in Occluded Environments}

\author{Yifei Chen$^{1}$, Kunyu Peng$^{1,*}$, Alina Roitberg$^{2}$, David Schneider$^{1}$, Jiaming Zhang$^{1}$, Junwei Zheng$^{1}$,\\Yufan Chen$^{1}$, Ruiping Liu$^{1}$, Kailun Yang$^{3,4}$, and Rainer Stiefelhagen$^{1}$%
\thanks{The project served to prepare the SFB 1574 Circular Factory for the Perpetual Product (project ID: 471687386), approved by the German Research Foundation (DFG, German Research Foundation). This work was supported in part by the SmartAge project sponsored by the Carl Zeiss Stiftung (P2019-01-003; 2021-2026), the MWK through the Cooperative Graduate School Accessibility through AI-based Assistive Technology (KATE) under Grant BW6-03, and in part by the BMBF through a fellowship within the IFI program of the German Academic Exchange Service (DAAD), in part by the HoreKA@KIT supercomputer partition, and in part by the National Natural Science Foundation of China (No. 62473139).}
\thanks{*Corresponding author. (Email: kunyu.peng@kit.edu.)}%
\thanks{$^{1}$The authors are the Institute for Anthropomatics and Robotics, Karlsruhe Institute of Technology, Germany.}%
\thanks{$^{2}$The author is with the Institute for Artificial Intelligence, University of Stuttgart, Germany.}%
\thanks{$^{3}$The author is with the School of Robotics, Hunan University, China.}%
\thanks{$^{4}$The author is also with the National Engineering Research Center of Robot Visual Perception and Control Technology, Hunan University, China.}
}

\maketitle
%\IEEEpeerreviewmaketitle 
\begin{abstract}
To integrate action recognition into autonomous robotic systems, it is essential to address challenges such as person occlusions—a common yet often overlooked scenario in existing self-supervised skeleton-based action recognition methods. In this work, we propose IosPSTL, a simple and effective self-supervised learning framework designed to handle occlusions. IosPSTL combines a cluster-agnostic KNN imputer with an Occluded Partial Spatio-Temporal Learning (OPSTL) strategy. First, we pre-train the model on occluded skeleton sequences. Then, we introduce a cluster-agnostic KNN imputer that performs semantic grouping using k-means clustering on sequence embeddings. It imputes missing skeleton data by applying K-Nearest Neighbors in the latent space, leveraging nearby sample representations to restore occluded joints. This imputation generates more complete skeleton sequences, which significantly benefits downstream self-supervised models. To further enhance learning, the OPSTL module incorporates Adaptive Spatial Masking (ASM) to make better use of intact, high-quality skeleton sequences during training. Our method achieves state-of-the-art performance on the occluded versions of the NTU-60 and NTU-120 datasets, demonstrating its robustness and effectiveness under challenging conditions. Code is available at \url{https://github.com/cyfml/OPSTL}.
\end{abstract}


\section{Introduction}
Human action recognition has extensive applications in the field of robotics, such as human-robot interaction, healthcare, industrial automation, security, and surveillance~\cite{bandi2021skeleton, Human-Robot-Interaction, healthcare,industrial-automation, danafar2007action}. 
In particular, robots can collaborate with humans as partners and assist them in various tasks by identifying human actions and needs. 
The capability to understand human intentions and goals allows a robot to discern when its assistance is most needed, thereby minimizing disruptions to human activities. A robot equipped with a human action recognition system can also be used to monitor the condition of patients to provide better daily-life assistance for their recovery, assess the safety of its surroundings, issue warnings, and detect gestures for help in rescue missions to provide assistance.
Challenges of image- or video-based action recognition~\cite{shaikh2021rgb} stem from multiple factors: complex backgrounds, variations in human body shapes, changing viewpoints, or motion speed alterations. In contrast to video-based action recognition~\cite{ullah2017action,pham2022video}, skeleton-based action recognition is less sensitive to appearance factors and has the advantage of superior efficiency by using sparse 3D skeleton data as input, which ensures fast inference speed and small memory usage. Thanks to the advancement of depth sensors~\cite{KinectSensor} and lightweight and robust pose estimation algorithms~\cite{cao2017realtime, PoseEstimation}, obtaining high-quality skeleton data is becoming easier.
Skeleton-based action recognition has rapidly progressed in recent years. Its efficiency designates it for mobile robots with computational constraints; at the same time, self-supervised solutions~\cite{liu2021self,kahn2018self} have gradually grasped the attention of the robot research community since this technique allows for training such methods with little annotation effort. 
\begin{figure}[t]
    \centering
    \begin{minipage}{.70\linewidth}
            \begin{subfigure}[t]{.99\linewidth}
                \includegraphics[width=\textwidth]{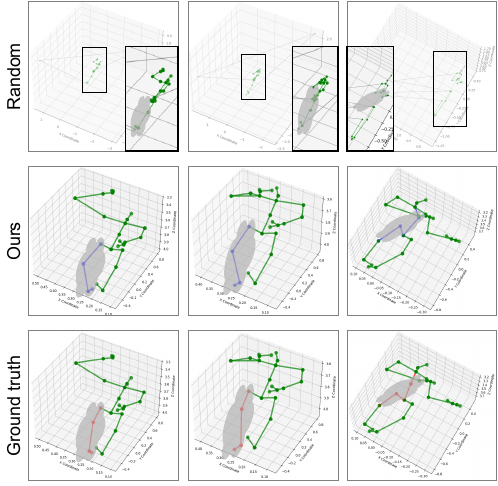}
                \caption{Comparison of imputations under occlusion
                }
                \label{fig:occl_comp}
            \end{subfigure}
        \end{minipage}
    \begin{minipage}{.26\linewidth}
        \begin{subfigure}[t]{.99\linewidth}
            \includegraphics[width=\textwidth]{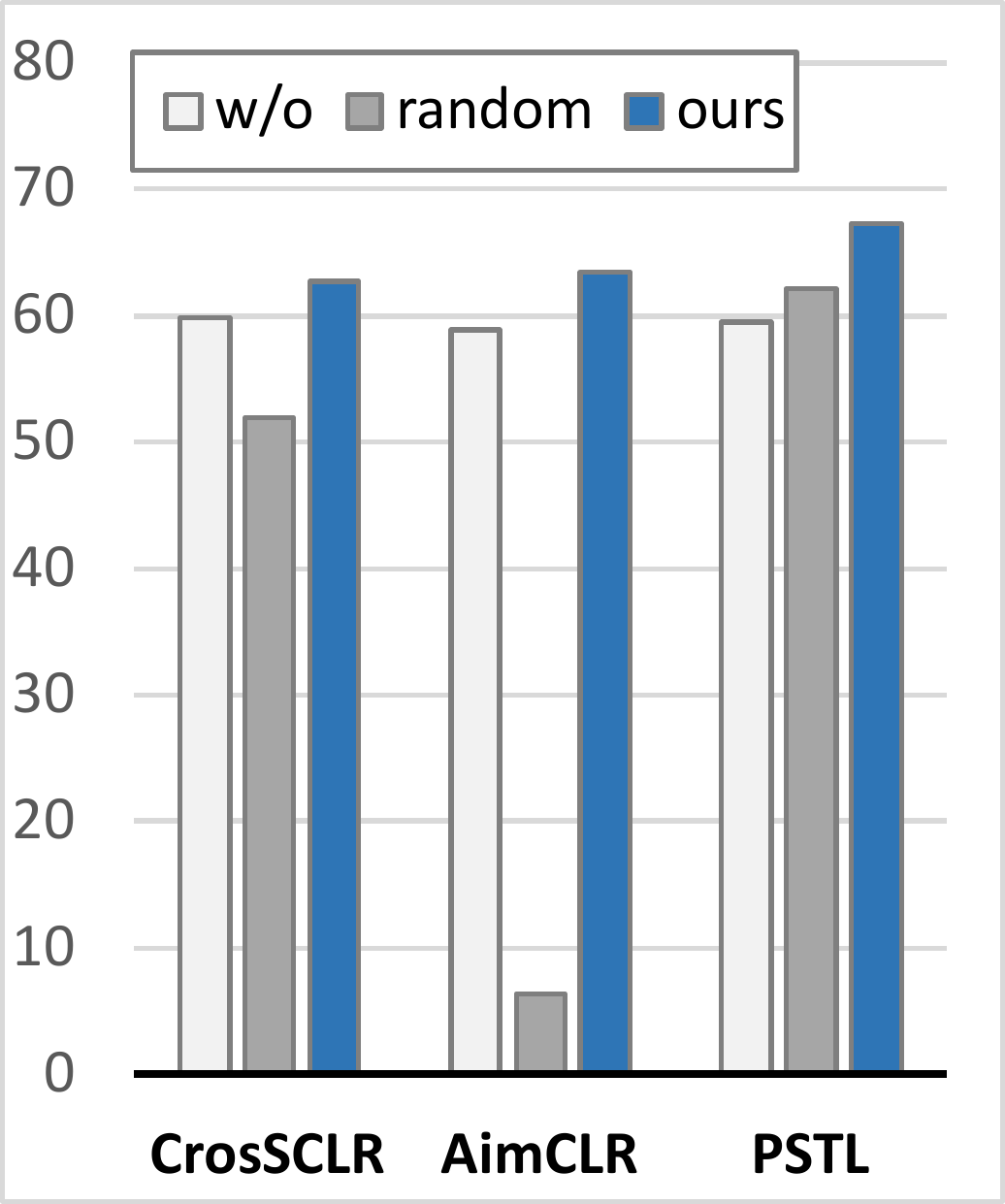}
            \caption{xsub}
            \label{fig:res_xsub}
        \end{subfigure} \\
        \begin{subfigure}[b]{.99\linewidth}
            \includegraphics[width=\textwidth]{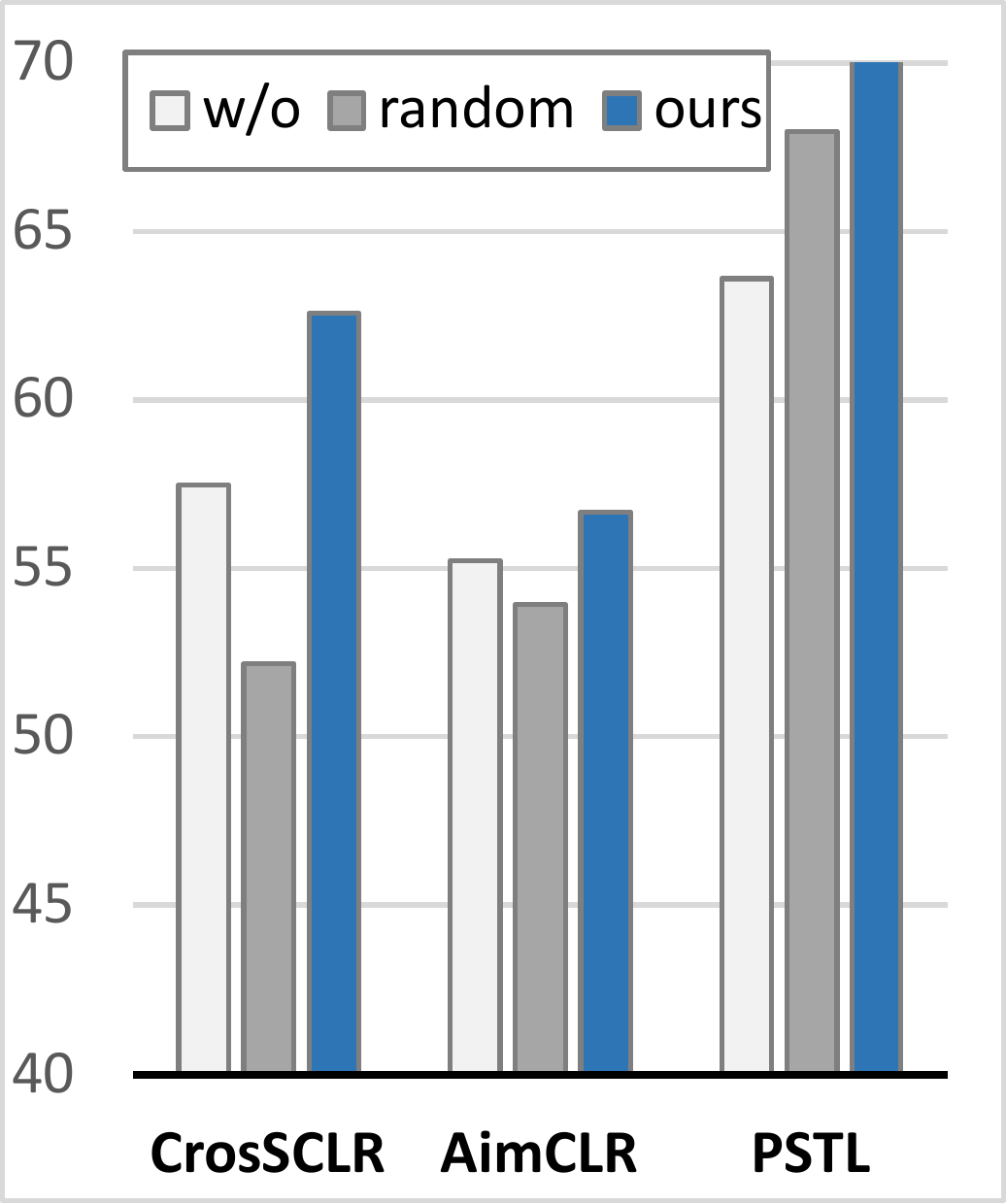}
            \caption{xview}
            \label{fig:res_xview}
        \end{subfigure} 
    \end{minipage}
    \caption{Comparison of different imputation methods. In (a), we compare random imputations (in \textcolor{gray}{gray}), our imputation results (in \textcolor{blue}{blue}), and ground-truth skeletons (in \textcolor{red}{red}). In (b) and (c), the linear evaluation results of cross-subject (\textit{xsub}) and cross-view (\textit{xview}) settings are tested by using imputation methods across three popular self-supervised action recognition methods (CrossCLR~\cite{li20213d}, AimCLR~\cite{guo2021contrastive}, and PSTL~\cite{zhou2023selfsupervised}).}
    \vskip-5ex
    \label{fig:imputed_joints_visiualization}
\end{figure}
The majority of existing work on self-supervised skeleton-based action recognition~\cite{Lin2020MS2LMS,li20213d,guo2021contrastive} is conducted on occlusion-free data collected in well-constrained environments. In practice, robots often encounter occluded environments in the real world, even high-quality pose detectors can not provide reliable full-body poses in such situations. For this reason, we argue that occlusion-aware self-supervised skeleton based human action recognition is an overlooked but crucial task in this field.
The occlusion problem in self-supervised skeleton-based action recognition can be considered from two points of view. On one hand, it can be addressed by improving the robustness of the model to occlusion by manipulating the model architecture, and on the other hand, it can be handled through the data itself by completing the missing skeleton coordinates as much as possible.
In this work, we for the first time tackle the self-supervised skeleton-based action recognition task under occlusions. Due to the lack of research in this field, we contribute the first benchmark on the NTU60 and NTU120 datasets with the realistic synthesized occlusion derived from Peng~\textit{et al.}~\cite{peng2023delving} and use well-established self-supervised skeleton based human action recognition approaches as baselines. The realistic synthesized occlusion is generated by projecting 3D IKEA furniture models into the skeletons' 3D coordinate space and applying ray casting from the camera center to each body joint to determine whether a joint is occluded.
On this benchmark, obvious performance decays of the utilized baselines are observed when using occluded skeleton data. A robust approach is needed by the community to achieve more robust self-supervised skeleton-based human action recognition.
We thereby contribute a new method, IosPSTL, by considering both model and data perspectives to impute occluded skeleton sequences, and then evaluate the performance of imputed skeleton sequences. 

From the model perspective, we introduce a novel dataset-driven \emph{Adaptive Spatial Masking (ASM)} data augmentation to enhance the robustness of the model toward occlusion perturbation.
This method masks joints based on the distribution of missing joints within the dataset to effectively leverage intact data to learn feature representation.
For the data-driven approach, we propose \emph{cluster agnostic KNN imputer} that further enhance the performance of self-supervised skeleton-based human action recognition, as shown in Fig.~\ref{fig:imputed_joints_visiualization}. The cluster agnostic KNN imputer can generalize to other self-supervised skeleton-based human action recognition baselines. 
Intuitively, one might search for similar dataset samples to fill in missing data. However, due to the vast amount of data and the density of the original skeleton data, directly applying KNN~\cite{KNN} to search for neighboring samples is highly impractical and unacceptable in terms of both time and space considerations. 
To improve computational efficiency, we propose a two-stage approach. In the first stage, samples are grouped into distinct categories through KMeans~\cite{kmeans} clustering on features learned through self-supervised learning methods. In the second stage, missing values are imputed by leveraging close neighbors within the same cluster.

The proposed approach eliminates the need for a KNN search on the entire dataset during the imputation process. Instead, cluster agnostic KNN imputation is applied within each smaller cluster, leading to a considerable reduction in computational overhead. We summarize our contributions as follows:
\begin{itemize}
  \item To investigate robotic action recognition performance in difficult environments, we construct the first large-scale occlusion-based benchmark for self-supervised skeleton-based action recognition, including both NTU-60 and NTU-120 datasets.  
  \item We propose a two-stage imputation method, named \emph{cluster agnostic KNN imputer}, using KMeans and KNN to reduce computation overhead for the occluded skeleton completion. Our imputed skeleton sequences show a huge improvement over the non-imputed skeleton sequence. It's also flexible and applicable to various self-supervised skeleton-based action recognition methods.
  \item We present the \emph{Occluded Partial Spatio-Temporal Learning (OPSTL)} framework, which leverages high-quality skeleton data using dataset-driven \emph{Adaptive Spatial Masking (ASM)}. Extensive experiments on occluded NTU-60 and NTU-120 datasets show that our method significantly improves accuracy, achieving a gain of about $7\%$ with realistic synthesized occlusions.

\end{itemize}

\section{Related Work}
\subsection{Skeleton-based Action Recognition} Early existing approaches for skeleton-based action recognition primarily concentrated on developing hand-crafted features. 
With the rapid advancement of deep learning,
early deep-learning methods included transforming skeleton data into images and utilizing convolutional neural networks (CNNs)~\cite{CNN1, CNN2, CNN3} for resolution, as well as directly utilizing recurrent neural networks (RNNs)~\cite{RNN1, RNN2, RNN3} to process skeleton data. ST-GCN~\cite{STGCN} initially proposed to treat skeleton data as pre-defined graphs and use graph convolutional neural networks (GCN) to aggregate information between joints. Subsequently, various methods based on ST-GCN have been continuously introduced, such as GCNs with attention mechanisms and multi-stream GCNs~\cite{2sAGCN,si2019attention,chen2022multiscale}. In this paper, existing popular methods all adopt ST-GCN as the backbone for feature extraction.

\subsection{Self-supervised Representation Learning} In the early stages of self-supervised learning, novel pretext tasks were designed to generate supervision from the inherent characteristics of the data itself, \textit{e.g.}, jigsaw puzzles, colorization, and predicting rotation. However, their performance heavily relies on the design of pretext tasks, and the generalization performance to downstream tasks cannot be guaranteed. Then, Instance discrimination-based contrastive learning methods, \textit{e.g.}, MOCO and MOCOv2~\cite{he2020momentum,chen2020improved}, utilize queue-based memory banks to store a large number of negative samples and employ momentum updating mechanisms. Additionally, SimCLR~\cite{chen2020simple} computes embeddings in real time using larger batch sizes. They all require a substantial number of negative samples for contrastive learning. Therefore, negative-sample-free methods like BYOL~\cite{grill2020bootstrap}, SimSiam~\cite{chen2020exploring}, and Barlow Twins~\cite{zbontar2021barlow} have been proposed to break free from the constraint of requiring a large number of negative samples for contrastive learning. The majority of these methods employ an asymmetric network architecture to prevent feature collapse. Notably, Barlow Twins circumvents the need for a symmetry-breaking network and instead reduces redundancy within the representation vector to mitigate collapse. 
Recently, MAE~\cite{he2021masked} has been introduced for learning informative visual representations through the utilization of local complementary information.

\subsection{Self-supervised Skeleton-based Action Recognition} 
MS2L~\cite{Lin2020MS2LMS} introduces a multi-task self-supervised learning framework involving motion predictions and jigsaw puzzles.
SkeletonCLR~\cite{li20213d} utilizes momentum updates in contrastive learning on individual streams. However, CrosSCLR~\cite{li20213d} goes beyond single-stream considerations. It employs a cross-view knowledge mining strategy to facilitate knowledge sharing between different streams, aiming to extract more valuable information. AimCLR~\cite{guo2021contrastive} recognizes the significance of data augmentation in contrastive learning and thus explores a multitude of data augmentation methods and combines them. 
On the other hand, PSTL~\cite{zhou2023selfsupervised} argues that these contrastive learning methods overly rely on data augmentation and don't consider the redundancy of spatial joints and temporal frames. Therefore it uses a spatiotemporal masking strategy to learn more generalized representations from partial skeleton sequences.

In this work, we first construct a large-scale self-supervised skeleton-based action recognition benchmark considering existing well-established self-supervised skeleton-based action recognition approaches under realistic occlusion from~\cite{peng2023delving} and then propose an \emph{Imputation Occluded Skeleton from Partial Spatial-Temporal Learning (IosPSTL)} framework by using OPSTL and \emph{cluster agnostic KNN imputer}, simultaneously.

\begin{figure*}[t]
  \centering
  \includegraphics[width=1\textwidth,
  ]{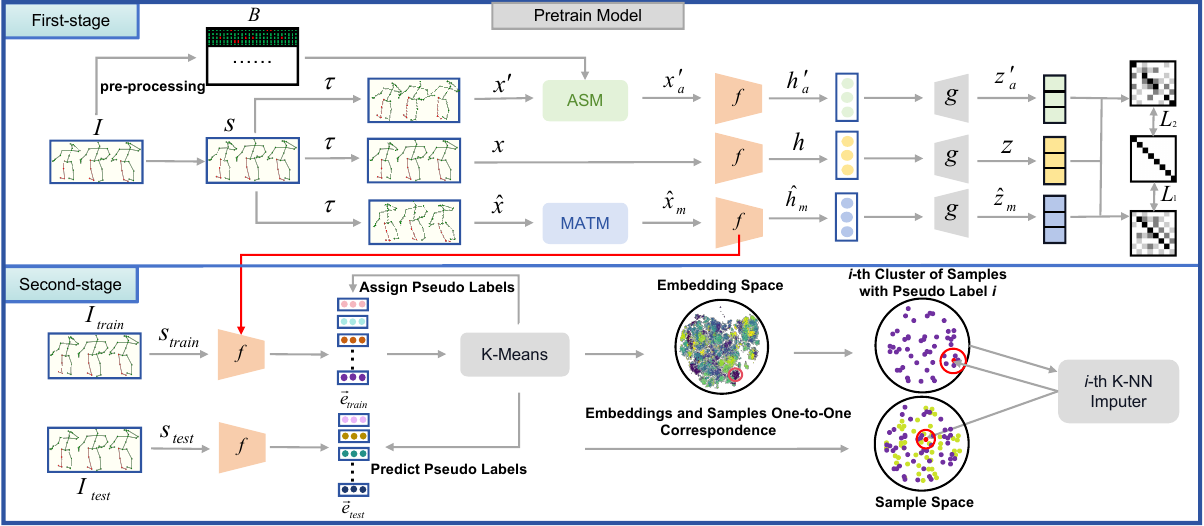}
  \caption{Our two-stage method for completing missing skeleton coordinates. The red portion in input $I$ represents the missing skeleton. In the first stage, the pre-training model adopts the PSTL framework, with CSM replaced by ASM, to better utilize high-quality data. The second stage involves completing the entire dataset by partitioning samples into smaller clusters using KMeans. Subsequently, the cluster-agnostic KNN-imputer is proposed to find neighboring samples and complete the missing coordinates. Yellow points in the sample space are samples from the test set, and purple points are samples from the training set.}
  \label{fig:pipeline}
  \vskip-3ex
\end{figure*}

\section{Methodology}
Our method, \textit{i.e.}, IosPSTL, is composed of a skeleton imputation method named cluster agnostic KNN-Imputer and a new self-supervised skeleton-based contrastive learning framework named OPSTL.
Occlusions are handled through the proposed cluster-agnostic KNN imputer.
We cluster the features obtained in the first stage of self-supervised training through KMeans. In each feature cluster, the cluster agnostic KNN imputer is used to find neighboring samples for imputation, as shown in Fig.~\ref{fig:pipeline}. After that, we obtain imputed skeleton sequences and train them the same way as in the first stage.

OPSTL is based on PSTL~\cite{zhou2023selfsupervised} due to its superior performance in comparison to other approaches when handling occlusions. The Central Spatial Masking (CSM) of PSTL is replaced by our proposed Adaptive Spatial Masking (ASM) to better handle occlusion in the first self-supervised training phase (first stage). 
Note that, the proposed cluster-agnostic KNN imputer in the second stage can be used on all self-supervised skeleton-based action recognition methods.
After imputing the missing data, downstream tasks can achieve improvements compared with occluded data using the same self-supervised skeleton-based action recognition method. 

\subsection{Pre-processing}

A pre-processed skeleton sequence can be represented as $s \in \mathbb{R}^{C \times T \times V}$ from the original input $I \in \mathbb{R}^{C \times T \times V \times M}$.  $T$ is the frame number and $V$ is the joint number. $C$ denotes the channel number, representing the 3D position. $M$ represents the person number. The preprocessing is similar to that of CrosSCLR~\cite{li20213d}. Skeleton coordinates are relative coordinates, relative to the center joint ($21$-th) of the skeleton. We find through experiments that relative coordinates are more robust to the training of occluded skeleton sequences because the invariance of relative coordinate representation makes it not affected by absolute position. Additionally, we need to compute the distribution of missing joints, represents a boolean matrix ($\mathbf{B} \in \mathbb{B}^{N \times V}$) of the $V$ joints for each sample to better mask the joints that are more occluded in the Adaptive Spatial Masking (ASM). 

\subsection{Partial Spatio-Temporal Skeleton Representation Learning}
Many existing methods~\cite{li20213d,guo2021contrastive} focus on generating various views of skeleton sequences for contrastive learning, but they often overlook the local relationships between different skeleton joints and frames. However, these local relationships are vital for real-world applications because they provide critical context for tasks like action recognition. To bridge this gap, Partial Spatio-Temporal Learning (PSTL) leverages local relationships by using a unique spatiotemporal masking strategy to create partial skeleton sequences. 

This approach, known as Central Spatial Masking (CSM), involves masking certain joints in the spatial dimension, with a preference for joints exhibiting a higher degree of centrality, thus assigning them a higher probability of being masked. Additionally, it incorporates Motion Attention Temporal Masking (MATM), which involves masking specific frames in the temporal dimension. This is achieved by calculating motion values for each frame of the action, which act as attention weights to select the frames to be masked. 
These masking strategies force the encoder to focus on local relationships between joints and frames and generate similar features from the partial skeleton sequence compared to the whole skeleton sequence. Therefore, PSTL shows a certain level of effectiveness in handling occlusion. The skeleton sequences are utilized in a triplet stream structure comprising an anchor stream in the middle, a spatial masking stream with CSM, and a temporal masking stream with MATM, as shown in Fig.~\ref{fig:pipeline}. In the first stage, we adopt the framework of PSTL, the difference is that ASM is used instead of CSM. In addition, the input of PSTL is a complete skeleton sequence, and our input is a skeleton sequence with occlusion. Each stream shares one encoder $f$ and one projector $g$.

PSTL adopts the framework of Barlow Twins~\cite{zbontar2021barlow}, thus avoiding the drawbacks of contrastive learning that require a large number of negative samples, as well as the need for a large batch size and memory bank~\cite{li20213d,guo2021contrastive}. By promoting the empirical cross-correlation matrix between embeddings of the distorted variations to be an identity matrix, the encoder can effectively capture the relationship between a masked stream and an anchor stream. The loss consists of two parts: one corresponds to the correlation matrix between the spatial stream and the anchor stream, while the other corresponds to the correlation matrix between the temporal stream and the anchor stream. The loss can be formulated as:
\begin{equation}
L = L_1+L_2,
\end{equation} where
\begin{equation}
L_1 = \sum_{i}(1 - {\hat{C}_{ii}})^2 + \lambda \sum_{i}\sum_{j \neq i} {\hat{C}_{ij}}^2
\end{equation} and
\begin{equation}
L_2 = \sum_{i}(1 - {C_{ii}^{'}})^2 + \lambda \sum_{i}\sum_{j \neq i} {C_{ij}^{'}}^2.
\end{equation}

Here, $\lambda$ acts as a trade-off parameter, balancing between the two terms. $C$ is the cross-correlation matrix computed between
the anchor embedding $z$ and one masked stream embedding $z^{'}$ and along the batch dimension b:
\begin{equation}
C_{ij} = \frac{\sum\nolimits_b{z_{b,i}z_{b,j}^{'}}}{\sqrt{\sum\nolimits_b{(z_{b,i})^{2}}}\sqrt{\sum\nolimits_b{(z_{b,j}^{'})^{2}}}},
\end{equation}
where $i$ and $j$ represent the embedding dimension.

\subsection{Adaptive Spatial Masking}

The self-supervised skeleton-based action recognition method, PSTL, employs Central Spatial Masking (CSM) to enhance the robustness of the learned representation with respect to joints. CSM promotes generating similar features from partial skeleton data and whole skeleton data, enabling the encoder to learn the relationship between masked and unmasked joints. CSM selects joints to be masked based on the degree of centrality of the human skeleton graph topology because the joints with more degrees can acquire richer neighborhood information. The joints with higher degrees are more likely to be masked. Joints' masked probability is defined as:
\begin{equation}
p_i = \frac{d_i}{\sum_{j=1}^{n} d_j},
\label{eq:CSM}
\end{equation}
$d_i$ is the degree of each joint $v_i$.

However, it does not take into account the actual occlusion situation. When only some joints of each sample are occluded, we can choose to mask the joints with higher occlusion frequency in the training set to better utilize high-quality data for learning.
On the other hand, when each sample is randomly occluded with a higher occlusion rate, the masking strategy should also shift from a fixed strategy to a random mask, in order to better simulate the distribution of occlusions. Therefore, we propose dataset-driven Adaptive Spatial Masking (ASM), which can adaptively switch between partial occlusion and random occlusion. It is worth noting that we still retain CSM in ASM. When no joints are occluded in this batch, we still use CSM to select joints for masking. We redefine the degree of each joint based on the missing joint boolean matrix ($\mathbf{B}$) in each batch. The missing frequency of each joint $v_i$, $i {\in} (1, 2, ..., n)$ is calculated within each batch. The frequency degree (FD) of joint $v_i$ can be formulated as:
\begin{equation}
    FD_i = \lfloor \frac{F_{i} - \min(F)}{\max(F) - \min(F) + \epsilon} \times 3 + 1 \rfloor,
    \label{eq:ASM1}
\end{equation}
where $\epsilon$ is a small value of $0.001$.
We observed that the majority of joints have a degree of around $2$, and the differences in degrees are relatively small. Consequently, the performance difference between random masking and degree-based masking is not significant. Thus, we rescale the frequency of each joint's occlusion to a range similar to the degrees, \textit{i.e.}, $[1,3]$. Due to the larger differences in frequency degrees generated by this rescaling compared to the degree of centrality of the human skeleton graph topology, the joint masking tends to favor joints with a higher frequency of occlusion. Here $F$ is the frequency of missing joints computed on $\mathbf{B}$ along the batch dimension $b$:
\begin{equation}
    F_i = \sum\nolimits_b \mathbf{B}_{b,i}.
    \label{eq:ASM}
\end{equation}

\subsection{Cluster Agnostic KNN Imputer}
We aim to find the most similar samples to the ones with missing values for imputation. However, due to the high dimensionality and large amount of sample data, directly searching for neighbors in the sample space is impractical. Therefore, we consider it unnecessary to search the entire sample space. Instead, we divide the samples into clusters with fewer skeleton samples. Through the first stage of pre-training, KMeans can roughly cluster samples with the same action type into a cluster. 

Firstly, we need to use the pre-trained model from the first stage to extract features from the samples. The extracted embeddings $\vec{e}_{train} \in  \mathbf{R}^{N \times D}$ are clustered using KMeans, where $N$ is the number of samples in the training set and $D$ is the dimension of embeddings. Pseudo-labels are assigned to each embedding. Since there is a one-to-one correspondence between embeddings and original samples, each sample is also assigned a pseudo-label. For a given cluster of samples with pseudo label $i$, we utilize KNN to search for neighboring samples of the sample that needs imputation within the same cluster. Because these neighboring samples may also have missing values, the standard Euclidean distance is not applicable. Here, we use a modified Euclidean distance based on missing values \cite{dixon1979pattern,scikit-learn}, which is formulated as:
\begin{equation}
    dist(S_{ij}, S_{ik}) = \sqrt{w \times d_{ignore}(S_{ij}, S_{ik})},
\end{equation}
where $w$ is a weight that can be expressed as the ratio of the total number of coordinates to the number of present coordinates, and $d_{ignore}(S_{ij},S_{ik})$ is the Euclidean distance between sample $j$ and sample $k$ in $i$-th cluster that ignores missing values in $S_{ij}$ and $S_{ik}$.

Based on this distance metric, it is straightforward to compute distances between each pair of samples. The nearest $k$ samples $S_{ij}^{\,near}$, $j \in (1, 2, ..., k)$ are selected based on distance and missing position from the current cluster $i$ for one of the samples with missing data $S^{\,miss}_{im}$, $m\neq j$, and each sample $S_{ij}^{\,near}$ should have intact coordinate $c_j \in C_j$ at positions $P  \in \mathbb{Z}^{T \times V \times M}$ where missing coordinates $C_{m} = \{c \mid p \in P \text{ and }(c,p) \in S_{im}^{\,miss} \text{ and } c \text{ is missing} \}$ occur. If there are missing values in the corresponding position $p \in P$ of the k-nearest samples $S_{ij}^{\,near}$ that need to be imputed, the KNN-imputer will look for the next nearest neighboring sample where the corresponding position $p$ is not missing. Therefore, these k-nearest samples are not fixed.  
The imputation formula for a missing skeleton coordinate of a missing sample $S_{im}^{\,miss}$ at a position $p$ is given as:
\begin{equation}
    c \in C_m =  \frac{\sum_{j=1}^{k}r_j \times c_j}{\sum_{j=1}^{k}r_j}, m\neq j
\end{equation}
where $r_j$ is the reciprocal of the modified Euclidean distance, denoted by $dist$, between a missing sample $S_{im}^{\,miss}$ and one of the nearest $k$ samples $S_{ij}^{\,near}$:
\begin{equation}
    r_j =  \frac{1}{dist(S_{ij}^{\,near},S_{im}^{\,miss})}.
\end{equation}

As shown in Fig.~\ref{fig:pipeline}, the difference between the imputation of the training set and the test set is that we don't recluster the test set from scratch.
Instead, the KMeans model trained on the training set is used to predict pseudo-labels for the test set. The imputed data is then generated using clusters from the training set that share the same pseudo-labels as those predicted for the test set. The test set is solely utilized as a source of data requiring imputation.

While this approach has shown improvements in various downstream tasks across multiple models, there are still limitations. When certain joints of all data in a cluster are missing, the missing parts cannot be imputed. Although the likelihood of this happening is extremely small, it does not guarantee the complete imputation of all missing skeleton coordinates.

\section{Experiments}

\subsection{Datasets}

\noindent\textbf{NTU-RGB+D 60/120 with occlusion.} The occluded datasets are derived from NTU-60/120. The NTU-60 dataset~\cite{shahroudy2016ntu} was captured by using Microsoft Kinect sensors and comprises $56,578$ skeleton sequences involving $60$ distinct action categories. There are two splits~\cite{shahroudy2016ntu}: 1) Cross-Subject (xsub): training data and validation data are captured from different subjects. 2) Cross-View (xview): training data and validation data are captured from different camera views. NTU-120 dataset \cite{ntu-120} is the extended version of the NTU-60, which comprises $113,945$ skeleton sequences involving $120$ action categories. NTU-120 keeps the xsub protocol while using the xset protocol to evaluate with different camera setups instead of views.
There are two types of occlusions: 1) Synthesized realistic occlusion~\cite{peng2023delving}, which leverages 3D furniture projections to generate realistic occlusions. Note that we only employ the dataset proposed by Peng~\textit{et al.}~\cite{peng2023delving} since their work focuses on one-shot skeleton-based human action recognition, which is a different task compared with self-supervised skeleton-based human action recognition.
The realistic occlusion involves projecting 3D furniture models onto 3D skeletons using different geometric parameters such as rotation and displacement.
2) Random occlusion, which is according to the minimum and maximum values of the coordinates, where $20\%$ of the coordinates are randomly selected for occlusion. 
\begin{table}[t]
\caption{Linear evaluation results on \textbf{NTU-60} and \textbf{NTU-120} with synthesized realistic occlusion. Imputation Occluded Skeleton from Partial Spatial-Temporal Learning (IosPSTL) combines OPSTL and Imputation methods. \textbf{J} and \textbf{M} indicate joints modality and velocity modality, respectively.}
\centering
%\setlength{\tabcolsep}{2px}
%\resizebox{\columnwidth}{!}{
\begin{tabular}{c|c|cc|cc}
\midrule
\multirow{2}{*}{\centering\textbf{Method}} & \multirow{2}{*}{\textbf{Stream}} &\multicolumn{2}{c|}{\centering NTU-60} &\multicolumn{2}{c}{\centering NTU-120} \\
& & \multirow{1}{*}{\centering xsub} & \multirow{1}{*}{\centering xview} & \multicolumn{1}{c}{\centering xsub} &\multicolumn{1}{c}{\centering xview}  \\
\midrule
\midrule
SkeletonCLR\cite{li20213d} & \textbf{J} & 56.74 & 53.25 & 44.93& 42.78\\
2s-CrosSCLR\cite{li20213d} & \textbf{J+M} & 59.88 & 57.47 
 &49.63 &48.14\\
AimCLR\cite{guo2021contrastive} & \textbf{J} & 58.90 & 55.21 & 44.58& 48.93\\
PSTL\cite{zhou2023selfsupervised} & \textbf{J} & 59.52 & 63.60 & 54.18& 51.90\\
\hline
\textbf{IosPSTL(Ours)} & \textbf{J} & \textbf{67.11}  &\textbf{71.39}&\textbf{59.29} & \textbf{58.25}\\
\bottomrule
\end{tabular}
%}
\vskip-3ex
\label{tab:main_tab}
\end{table}
\begin{table*}[t]

\caption{Comparison among \textbf{no imputation}, \textbf{random imputation}, and our proposed \textbf{cluster agnostic KNN imputation} on various of approaches. Linear evaluation results on \textbf{NTU-60} with realistic synthesized occlusion are reported.``$\Delta$'' represents the difference compared to the non-imputed NTU-60. \textbf{J} and \textbf{M} represent the joint stream and the motion stream. 
Note that, our contributed method incorporates both the OPSTL and the new imputation method.}
\centering
%\setlength{\tabcolsep}{2px}
%\resizebox{\columnwidth}{!}{
\begin{tabular}{c|c|cc|cc|cc|cc|cc}
\midrule
\multirow{3}{*}{\centering\textbf{Method}} & \multirow{3}{*}{\textbf{Stream}} & \multicolumn{2}{p{1.8cm}|}{\centering\textbf{No imputation} (\%)} & \multicolumn{4}{p{3cm}|}{\centering\textbf{Random imputation} (\%)} & \multicolumn{4}{p{3cm}}{\centering\multirow{1}{*}{\textbf{Our imputed} (\%)}}\\
\cline{3-12}
& & \multirow{1}{*}{\centering xsub} & \multirow{1}{*}{\centering xview} &  \multicolumn{2}{c|}{\centering xsub} &\multicolumn{2}{c|}{\centering xview} & \multicolumn{2}{c|}{\centering xsub} & \multicolumn{2}{c}{\centering xview} \\
& & acc.  & acc.  & acc. & $\Delta$ & acc.  & $\Delta$& acc. & $\Delta$ & acc.  & $\Delta$\\
\midrule
\midrule
SkeletonCLR~\cite{li20213d} & \textbf{J} & 56.74 & 53.25 & 47.12 & \deltdown{9.62} &58.09& \deltup{4.84}& \cellcolor{blue!25}57.61 & \cellcolor{blue!25}\deltup{0.87} &\cellcolor{blue!25}64.43&\cellcolor{blue!25} \deltup{11.18}\\
2s-CrosSCLR~\cite{li20213d} & \textbf{J+M} & 59.88 & 57.47 & 51.96 & \deltdown{7.92} &52.18 & \deltdown{5.29}& \cellcolor{blue!25}62.76 & \cellcolor{blue!25}\deltup{2.88} &\cellcolor{blue!25}62.54& \cellcolor{blue!25}\deltup{5.07}\\
AimCLR~\cite{guo2021contrastive} & \textbf{J} & 58.90 & 55.21 & 6.36 & \deltdown{52.54} &53.91& \deltdown{1.30}& \cellcolor{blue!25}63.40 & \cellcolor{blue!25}\deltup{4.50} &\cellcolor{blue!25}56.68& \cellcolor{blue!25}\deltup{1.47}\\
PSTL~\cite{zhou2023selfsupervised} & \textbf{J} & 59.52 & 63.60 & 62.18 & \deltup{2.66}&67.97&\deltup{4.37}& \cellcolor{blue!25}\textbf{67.31} & \cellcolor{blue!25}\deltup{7.79} &\cellcolor{blue!25}71.10& \cellcolor{blue!25}\deltup{7.50} \\
OPSTL (ours)& \textbf{J} & \textbf{61.11} & \textbf{65.55} & \textbf{65.63} &\deltup{4.52} &\textbf{68.01}&\deltup{2.46}& \cellcolor{blue!25}67.11 & \cellcolor{blue!25}\deltup{6.00} &\cellcolor{blue!25}\textbf{71.39}& \cellcolor{blue!25}\deltup{5.84} \\
\bottomrule
\end{tabular}
%}
\label{tab:non-imputed and randomly imputed ntu60}
\end{table*}

\begin{table*}[t]
\caption{Comparison among \textbf{no imputation}, \textbf{random imputation}, and our proposed \textbf{cluster agnostic KNN imputation} on various of approaches. Linear evaluation results on \textbf{NTU-120} with synthesized realistic occlusion are reported. ``$\Delta$'' represents the difference compared to the non-imputed NTU-120. \textbf{J} and \textbf{M} represent the joint stream and the motion stream.}
\centering
%\resizebox{\columnwidth}{!}{
%\setlength{\tabcolsep}{2px}
\begin{tabular}{c|c|cc|cc|cc|cc|cc}
\midrule
\multirow{3}{*}{\textbf{Method}} & \multirow{3}{*}{\textbf{Stream}} & \multicolumn{2}{p{1.8cm}|}{\centering\textbf{No imputation} (\%)} & \multicolumn{4}{p{3cm}|}{\centering\textbf{Random imputation} (\%)} & \multicolumn{4}{p{3cm}}{\centering\multirow{1}{*}{\textbf{Our imputation} (\%)}}\\
\cline{3-12}
& & \multirow{1}{*}{\centering xsub} & \multirow{1}{*}{\centering xset} &  \multicolumn{2}{c|}{\centering xsub} &\multicolumn{2}{c|}{\centering xset} & \multicolumn{2}{c|}{\centering xsub} & \multicolumn{2}{c}{\centering xset} \\
& &acc.  &acc.  & acc. & $\Delta$ & acc.  & $\Delta$& acc. & $\Delta$ & acc.  & $\Delta$\\
\midrule
\midrule
SkeletonCLR~\cite{li20213d} & \textbf{J} & 44.93 & 42.78 & 44.42 & \deltdown{0.51} &40.12& \deltdown{2.66}& \cellcolor{blue!25}48.63 & \cellcolor{blue!25}\deltup{3.70} &\cellcolor{blue!25}45.06& \cellcolor{blue!25}\deltup{2.28}\\
2s-CrosSCLR~\cite{li20213d} & \textbf{J+M} & 49.63 & 48.14 & 39.11 & \deltdown{10.52} &33.77 & \deltdown{14.37}&\cellcolor{blue!25} 49.58 &\cellcolor{blue!25} \deltdown{0.05} &\cellcolor{blue!25}54.43 & \cellcolor{blue!25}\deltup{6.29}\\
AimCLR~\cite{guo2021contrastive} & \textbf{J} & 44.58 & 48.93 & 0.86 & \deltdown{43.72} &1.16& \deltdown{47.77}& \cellcolor{blue!25}52.50 & \cellcolor{blue!25}\deltup{7.92} &\cellcolor{blue!25}52.83& \cellcolor{blue!25}\deltup{3.90}\\
PSTL~\cite{zhou2023selfsupervised} & \textbf{J} & 54.18 & 51.90 & 56.12 & \deltup{1.94}&52.66&\deltup{0.76} &\cellcolor{blue!25} 57.05 & \cellcolor{blue!25}\deltup{2.87}&\cellcolor{blue!25}57.94&\cellcolor{blue!25}\deltup{6.04}\\
\textbf{OPSTL (ours)}& \textbf{J} & \textbf{55.65} & \textbf{54.18} & \textbf{56.43} &\deltup{0.78} &\textbf{53.90}&\deltdown{0.28} &\cellcolor{blue!25}\textbf{59.29} &\cellcolor{blue!25}\deltup{3.64} &\cellcolor{blue!25}\textbf{58.25}&\cellcolor{blue!25}\deltup{4.07}\\
\bottomrule
\end{tabular}
%}
\label{tab:non-imputed and randomly imputed ntu120}
\end{table*}
\subsection{Protocols}
\noindent\textbf{Linear Evaluation Protocol}. 
To elaborate, we train a supervised linear classifier consisting of a fully connected layer followed by a softmax layer while keeping the encoder fixed.

\noindent\textbf{Semi-supervised Evaluation Protocol}. We initially pre-train the encoder using the entire imputed dataset and subsequently fine-tune the complete model using only $1\%$ or $10\%$ randomly chosen labeled data.

\noindent\textbf{Finetune Protocol}. We attach a linear classifier to the trained encoder and finetune the entire network on the imputed data.

\subsection{Implementation Details}
In our experiments, all pre-trained models are based on ST-GCN~\cite{STGCN} with $16$ hidden channels as the backbone. The preprocessing steps closely follow those used in CrosSCLR and AimCLR. This involves removing invalid frames from skeleton sequences, resizing sequences to $50$ frames using linear interpolation, and transforming coordinates into relative coordinates. We also compute the distribution of missing joints ($\mathbf{B}$) in the data.

For training, we employ the Adam optimizer~\cite{kingma2014adam} and use the CosineAnnealing scheduler with a total of $150$ epochs for both representation learning and downstream tasks. Our batch size is set to $128$. The learning rate is set to $5e{-}3$.

\noindent\textbf{Data Augmentation.}
Data augmentation is performed to diversify skeleton sequences before feature extraction during model training. Each model uses its specific set of data augmentation methods. For instance, SkeletonCLR and CrosSCLR utilize one spatial augmentation (Shear) and one temporal augmentation (Crop).
AimCLR employs four spatial augmentations (Shear, Spatial Flip, Rotate, Axis Mask) and two temporal augmentations (Crop, Temporal Flip), along with two spatiotemporal augmentations (Gaussian Noise and Gaussian Blur). 
PSTL uses three spatial augmentations (Shear, Rotate, Spatial Flip) and one temporal augmentation (Crop).

\noindent\textbf{Self-supervised Pre-training.}
To ensure a direct comparison with PSTL, we use the same set of parameters. As shown in  Fig.~\ref{fig:pipeline}, the incomplete skeleton sequence $S$ through transformation $\tau$ has three different views $x$, $x'$, $\hat{x}$. $x'$ and $\hat{x}$ pass through ASM and MATM to create partial skeleton sequence $x'_{a}$, $\hat{x}_{m}$. ST-GCN is used as the shared backbone $f$ to extract $256$-dimensional features $h$, $h'_{a}$, $\hat{h}_{m}$, which are then projected to $6144$-dimensional embeddings $z$, $z'_{a}$, $\hat{z}_{m}$ through shared projector $g$. To capture the relationship between masked joints and unmasked ones, we compute the cross-correlation matrix between embeddings $z$ and $\hat{z}_{m}$, as well as $z$ and $z'_{a}$. $L1$ and $L2$ are calculated using the two cross-correlation matrices. The loss parameter $\lambda$ is set to $2e^{-4}$, and a warm-up strategy of $10$ training epochs is applied. Weight decay is set to $1e^{-5}$. For ASM, $9$ joints are masked, and for MATM, $10$ frames are masked.

\noindent\textbf{Imputation.}
We proposed an imputation method to deal with occlusion. For clustering during imputation, we employ KMeans with $60$ clusters for NTU-60 and $120$ clusters for NTU-120 with realistic occlusions. We use KNN with a value of $k$ set to $5$ to search for neighboring samples in the imputation process. The chosen cluster number corresponds to the number of the total class.

\subsection{Benchmark Analysis}
We propose Imputation Occluded Skeleton from Partial Spatial-Temporal Learning (IosPSTL) to address the occlusion challenge in self-supervised action recognition. As shown in Table~\ref{tab:main_tab}, our approach achieves significant improvements on occluded datasets compared with the selected baselines, with approximately $8\%$ and $6\%$ gains on the occluded NTU-60 and NTU-120, respectively, compared to other methods. 
Then we would like to examine the effectiveness of the proposed cluster-agnostic KNN imputer on other self-supervised skeleton-based human action recognition baselines. We compare all the baselines together with our proposed OPSTL w/o imputer, w/ random imputer, and w/ cluster agnostic KNN imputer on NTU-60/120 with realistic occlusion.
As shown in Tables~\ref{tab:non-imputed and randomly imputed ntu60},~\ref{tab:non-imputed and randomly imputed ntu120}, and~\ref{tab:LPFTSemi-imputed-ntu120}, with the help of the proposed cluster agnostic KNN imputer, almost all three downstream task performances of all the investigated methods have shown obvious improvements on both of the NTU-60 and NTU-120 datasets. 

The performance gains brought by our proposed imputer are demonstrated on the right hand side of Table~\ref{tab:non-imputed and randomly imputed ntu60} and marked by $\Delta$. 

OPSTL achieves a notable enhancement of $6\%$ and $5.84\%$ in xsub and xview of imputed NTU60, respectively. AimCLR demonstrates improvements of $7.92\%$ and $3.90\%$ in xsub and xset of imputed NTU120, respectively. The enhancement in accuracy after imputation is indicated by red upward arrows.

\begin{table}[t!]
\caption{Finetune and  Semi-supervised results on the imputed NTU-60/120 with synthesized realistic occlusion.``$\Delta$'' represents the difference compared to the non-imputed NTU-60/120 with synthesized realistic occlusion. \textbf{J} and \textbf{M} represent the joint stream and the motion stream.}
\centering
\resizebox{\columnwidth}{!}{
\setlength{\tabcolsep}{2px}
\begin{tabular}{c|c|cc|cc|cc|cc}
\midrule
\multirow{3}{*}{\textbf{Method}} & \multirow{3}{*}{\textbf{Stream}} & \multicolumn{4}{p{3cm}|}{\centering\textbf{Imputed NTU-60} (\%)}& \multicolumn{4}{p{3cm}}{\centering\textbf{Imputed NTU-120} (\%)} \\
\cline{3-10}
& & \multicolumn{2}{c|}{xsub} & \multicolumn{2}{c|}{xview} & \multicolumn{2}{c|}{xsub} & \multicolumn{2}{c}{xset} \\
& & acc. & $\Delta$ & acc.  & $\Delta$ & acc. & $\Delta$ & acc.  & $\Delta$\\
\midrule
\midrule
\textbf{Finetune}:&&\\
SkeletonCLR~\cite{li20213d} & \textbf{J}  & \cellcolor{blue!25}70.58 & \cellcolor{blue!25}\deltup{3.22} &\cellcolor{blue!25}80.76& \cellcolor{blue!25}\deltup{4.42} &\cellcolor{blue!25}63.17 & \cellcolor{blue!25}\deltup{4.06} &\cellcolor{blue!25}62.12& \cellcolor{blue!25}\deltup{1.20}\\
2s-CrosSCLR~\cite{li20213d} & \textbf{J+M} & \cellcolor{blue!25}72.94 & \cellcolor{blue!25}\deltup{1.32} &\cellcolor{blue!25}80.34& \cellcolor{blue!25}\deltup{0.09} & \cellcolor{blue!25}65.06 & \cellcolor{blue!25}\deltup{0.39} &\cellcolor{blue!25}67.45 & \cellcolor{blue!25}\deltup{2.43}\\
AimCLR~\cite{guo2021contrastive} & \textbf{J}  & \cellcolor{blue!25}70.53 & \cellcolor{blue!25}\deltup{0.44} &\cellcolor{blue!25}75.52& \cellcolor{blue!25}\deltdown{3.21} & \cellcolor{blue!25}67.08 & \cellcolor{blue!25}\deltup{5.25} &\cellcolor{blue!25}66.62& \cellcolor{blue!25}\deltup{1.91}\\
PSTL~\cite{zhou2023selfsupervised} & \textbf{J} & \cellcolor{blue!25}75.16 & \cellcolor{blue!25}\deltup{2.48} &\cellcolor{blue!25}85.24& \cellcolor{blue!25}\deltup{2.05} & \cellcolor{blue!25}69.10 & \cellcolor{blue!25}\deltup{1.20}&\cellcolor{blue!25}\textbf{69.42}&\cellcolor{blue!25}\deltup{2.71} \\
\textbf{OPSTL (ours)}& \textbf{J} & \cellcolor{blue!25}\textbf{75.43} & \cellcolor{blue!25}\deltup{2.41} &\cellcolor{blue!25}\textbf{86.01}& \cellcolor{blue!25}\deltup{1.92} & \cellcolor{blue!25}\textbf{70.89} &\cellcolor{blue!25}\deltup{2.21} &\cellcolor{blue!25}69.14&\cellcolor{blue!25}\deltup{1.89} \\
\midrule
\textbf{Semi 1\%}:&&\\
SkeletonCLR~\cite{li20213d} & \textbf{J} & \cellcolor{blue!25}31.99 & \cellcolor{blue!25}\deltup{13.47} &\cellcolor{blue!25}31.18& \cellcolor{blue!25}\deltup{10.16} & \cellcolor{blue!25}20.45 & \cellcolor{blue!25}\deltup{3.13} &\cellcolor{blue!25}16.24& \cellcolor{blue!25}\deltup{2.23}\\
2s-CrosSCLR~\cite{li20213d} & \textbf{J+M} & \cellcolor{blue!25}32.66 & \cellcolor{blue!25}\deltup{4.69} &\cellcolor{blue!25}31.18& \cellcolor{blue!25}\deltup{10.35} & \cellcolor{blue!25}19.38 & \cellcolor{blue!25}\deltup{0.21} &\cellcolor{blue!25}20.12 & \cellcolor{blue!25}\deltup{7.59}\\
AimCLR~\cite{guo2021contrastive} & \textbf{J}  & \cellcolor{blue!25}34.44 & \cellcolor{blue!25}\deltup{5.28} &\cellcolor{blue!25}27.04& \cellcolor{blue!25}\deltup{8.92} & \cellcolor{blue!25}22.59 & \cellcolor{blue!25}\deltup{6.13} &\cellcolor{blue!25}20.68& \cellcolor{blue!25}\deltup{5.38}\\
PSTL~\cite{zhou2023selfsupervised} & \textbf{J} & \cellcolor{blue!25}\textbf{40.81} & \cellcolor{blue!25}\deltup{7.99} &\cellcolor{blue!25}\textbf{39.61}& \cellcolor{blue!25}\deltup{13.06} & \cellcolor{blue!25}27.43 & \cellcolor{blue!25}\deltup{5.92}&\cellcolor{blue!25}\textbf{25.52}&\cellcolor{blue!25}\deltup{6.52} \\
\textbf{OPSTL (ours)}& \textbf{J} & \cellcolor{blue!25}40.07 & \cellcolor{blue!25}\deltup{6.48} &\cellcolor{blue!25}38.65& \cellcolor{blue!25}\deltup{10.76} &\cellcolor{blue!25}\textbf{27.90} &\cellcolor{blue!25}\deltup{4.69} &\cellcolor{blue!25}24.57&\cellcolor{blue!25}\deltup{4.71} \\
\midrule
\textbf{Semi 10\%}:&&\\
SkeletonCLR~\cite{li20213d} & \textbf{J} & \cellcolor{blue!25}55.97 & \cellcolor{blue!25}\deltup{2.98} &\cellcolor{blue!25}60.83& \cellcolor{blue!25}\deltup{9.37} & \cellcolor{blue!25}44.37 & \cellcolor{blue!25}\deltup{2.33} &\cellcolor{blue!25}42.68& \cellcolor{blue!25}\deltup{7.72}\\
2s-CrosSCLR~\cite{li20213d} & \textbf{J+M} & \cellcolor{blue!25}59.17 & \cellcolor{blue!25}\deltup{3.16} &\cellcolor{blue!25}59.01& \cellcolor{blue!25}\deltup{3.86} & \cellcolor{blue!25}46.89 & \cellcolor{blue!25}\deltup{2.07} &\cellcolor{blue!25}48.24 & \cellcolor{blue!25}\deltup{8.44}\\
AimCLR~\cite{guo2021contrastive} & \textbf{J}  & \cellcolor{blue!25}59.64 & \cellcolor{blue!25}\deltup{2.30} &\cellcolor{blue!25}54.34& \cellcolor{blue!25}\deltdown{0.36} & \cellcolor{blue!25}48.38 & \cellcolor{blue!25}\deltup{6.25} &\cellcolor{blue!25}48.96& \cellcolor{blue!25}\deltup{3.63}\\
PSTL~\cite{zhou2023selfsupervised} & \textbf{J} & \cellcolor{blue!25}63.04 & \cellcolor{blue!25}\deltup{4.41} &\cellcolor{blue!25}68.89& \cellcolor{blue!25}\deltup{7.54} & \cellcolor{blue!25}53.26 & \cellcolor{blue!25}\deltup{2.80}&\cellcolor{blue!25}53.42&\cellcolor{blue!25}\deltup{4.14} \\
\textbf{OPSTL (ours)}& \textbf{J}& \cellcolor{blue!25}\textbf{64.04} & \cellcolor{blue!25}\deltup{5.26} &\cellcolor{blue!25}\textbf{70.04}& \cellcolor{blue!25}\deltup{6.22} & \cellcolor{blue!25}\textbf{54.71} &\cellcolor{blue!25}\deltup{2.90} &\cellcolor{blue!25}\textbf{53.50}&\cellcolor{blue!25}\deltup{3.38} \\
\bottomrule
\end{tabular}}
\vskip-3ex
\label{tab:LPFTSemi-imputed-ntu120}
\end{table}

Additionally, we conduct a series of comparison experiments to evaluate OPSTL.
As shown in Tables \ref{tab:non-imputed and randomly imputed ntu60},~\ref{tab:non-imputed and randomly imputed ntu120}, and~\ref{tab:LPFTSemi-imputed-ntu120}, for all the non-imputed, randomly imputed, and imputed NTU-60/120, OPSTL outperforms the most performing baseline, \textit{i.e.}, PSTL on linear evaluation. The proposed ASM mechanism delivers performance improvements of $1.59\%$ and $1.95\%$ on xsub and xview of NTU-60 with realistic occlusion, respectively. In addition, OPSTL achieves a performance gain of $1.47\%$ and $2.28\%$ on xsub and xset of NTU-120 with realistic occlusion. 
Not only on the non-imputed dataset but also on xsub and xset of imputed NTU-120 achieves $2.24\%$ and $0.31\%$ improvements.

\subsection{Ablation Study}
We conduct ablation experiments to demonstrate the effectiveness of the proposed adaptive Spatial Masking (ASM) and the cluster agnostic KNN imputer method.

1) To validate the effectiveness of the imputation method, we perform random imputation on NTU-60/120 with realistic occlusion. 
As shown in the Table~\ref{tab:non-imputed and randomly imputed ntu60}, the cluster-agnostic KNN imputer achieves $4.50\%$ and $1.47\%$ performance improvements (shown on the right-hand side) on the baseline AimCLR when comparing with the AimCLR without imputation. However, random imputation can not achieve desired performance improvements, resulting in $52.54\%$ and $3.30$ performance decays. This observation indicates that the cluster agnostic KNN imputer can achieve better performance than random skeleton imputation.

The experiments also revealed that random imputation is beneficial for models trained using a partial skeleton sequence but detrimental for models trained using a complete skeleton sequence, leading to performance degradation. This is because randomly imputed skeleton data provides compensatory information that is useful for model learning from partial skeleton data. Therefore, it is necessary to conduct further research on random imputation. From Table~\ref{tab:non-imputed and randomly imputed ntu60} and Table~\ref{tab:non-imputed and randomly imputed ntu120}, it is obvious that our proposed cluster agnostic KNN imputer outperforms random imputation on linear evaluation. The performances of SkeletonCLR, 2s-CrosSCLR, and AimCLR all deteriorate under random imputation. 

\begin{table}[t!]
\setlength{\tabcolsep}{2px}
\caption{Stepwise ablation results on realistic occluded and imputed NTU-120. Method\textsuperscript{1+2} denotes two stages during the pre-training. \textbf{S1} is first stage and \textbf{S2} is second stage. All experiments are on the joint stream.}
\centering
\resizebox{\columnwidth}{!}{
\begin{tabular}{c|cc|cc}
\midrule
\multirow{3}{*}{\textbf{Method\textsuperscript{1+2}}} & \multicolumn{2}{p{2cm}|}{\centering\textbf{(S1) Occluded NTU-120} (\%)} & \multicolumn{2}{p{2cm}<{\centering}}{\centering\textbf{(S2) Imputed NTU-120} (\%)}\\
\cline{2-5}
& \hspace{0.2cm}xsub & xset & \hspace{0.2cm}xsub & xset\\
\midrule
\midrule
PSTL~\cite{zhou2023selfsupervised} + PSTL~\cite{zhou2023selfsupervised} &\hspace{0.2cm}54.18	&51.90	&\hspace{0.2cm}\cellcolor{blue!25}57.05	&\cellcolor{blue!25}57.94\\
OPSTL (ours) + PSTL~\cite{zhou2023selfsupervised} & \hspace{0.2cm}55.65	&54.18	&\hspace{0.2cm}\cellcolor{blue!25}58.70	&\cellcolor{blue!25}57.52\\
\textbf{OPSTL (ours) + OPSTL (ours)} & \hspace{0.2cm}\textbf{55.65}	&\textbf{54.18}	&\hspace{0.2cm}\cellcolor{blue!25}\textbf{59.29}	&\cellcolor{blue!25}\textbf{58.25}\\
\bottomrule
\end{tabular}}
\label{tab:stepwise}
\end{table}

2) To better illustrate the effectiveness of ASM, we conducted a stepwise ablation study on the NTU-120 dataset. As shown in Table~\ref{tab:stepwise}, during the first stage of pre-training, we observed that using ASM yields accuracy improvements of $1.47\%$ for xsub and $2.28\%$ for xset over CSM. Building upon the ASM-based first stage, in the second stage, both CSM and ASM are employed. The results indicate in the second stage, continuing to use ASM yields gains of $0.59\%$ for xsub and $0.73\%$ for xset compared to using CSM.

3) To demonstrate the effectiveness of our imputation method on random occlusions, we validate it using OPSTL on the NTU-60/120 datasets. As shown in Table~\ref{tab:random occlusion by PSTL}, after the imputation, OPSTL exhibits significant improvements across various splits of the NTU-60/120 datasets.

\begin{table}[t]
\setlength{\tabcolsep}{2px}
\caption{Linear evaluation results of OPSTL on non-imputed and imputed NTU-60/120 with random occlusion. All experiments are on the joint stream.}
\centering
\resizebox{\columnwidth}{!}{
\begin{tabular}{c|cc|cc}
\midrule
\multirow{3}{*}{\textbf{Method}} & \multicolumn{2}{p{2.5cm}|}{\centering\textbf{Randomly occluded NTU-60} (\%)} & \multicolumn{2}{p{2.5cm}}{\centering\textbf{Randomly occluded NTU-120} (\%)}\\
\cline{2-5}
& \hspace{0.3cm}xsub & xview &\hspace{0.3cm} xsub & xset\\
\midrule
\midrule
Non-Imputed & \hspace{0.3cm}41.90 & 30.16 & \hspace{0.3cm}10.32 &4.22\\
Imputed &\hspace{0.3cm}\cellcolor{blue!25}47.31 & \cellcolor{blue!25}55.29 &\hspace{0.3cm}\cellcolor{blue!25}23.54 & \cellcolor{blue!25}18.59\\
\bottomrule
\end{tabular}}
\label{tab:random occlusion by PSTL}
\end{table}

\section{Conclusion}
%%%%%%%%%%%%%%%%%%%%%%%%%%%%%%%%
In this paper, we propose effective solutions for the challenges of self-supervised skeleton-based action recognition in occluded environments. First, we construct a large-scale occluded self-supervised skeleton-based human action recognition benchmark considering well-established approaches. 
Due to the limited performances of the leveraged baselines, we propose a new approach, named IosPSTL, which delivered state-of-the-art performances on the NTU-60/120 datasets.
As part of the IosPSTL, cluster agnostic KNN imputer method using KMeans and KNN, reducing computation overhead for occluded skeleton completion is proposed. Apart from the data-driven approach, OPSTL with ASM is proposed to improve the robustness of feature learning during the contrastive training. Experimental results validate the efficacy of our approach across various self-supervised skeleton-based human action recognition approaches, empowering robots to perform robust action recognition in real-world occluded scenarios. 

\bibliographystyle{IEEEtran}
\bibliography{bib}

\end{document}